\documentclass[10pt]{article}
\usepackage[utf8]{inputenc}
\usepackage[T1]{fontenc}
\usepackage[margin=1in, left=1.25in, right=1.25in, top=1in, bottom=1in]{geometry}
\usepackage[linktoc=all, bookmarksdepth=3, pdfstartview=FitH]{hyperref}
\hypersetup{colorlinks=true, linkcolor=blue, filecolor=magenta, urlcolor=cyan,}
\usepackage{amsmath}
\usepackage{amsfonts}
\usepackage{amssymb}
\usepackage[version=4]{mhchem}
\usepackage{stmaryrd}
\usepackage{bbold}
\usepackage{graphicx}
\usepackage[export]{adjustbox}
\usepackage[dvipsnames]{xcolor}
\usepackage{float}  
\usepackage{caption}  
\usepackage{booktabs}
\usepackage{multirow}
\usepackage{xcolor}
\usepackage[normalem]{ulem} 
\newcommand{\revise}[1]{#1}
\newcommand{\del}[1]{}

\usepackage[ruled,vlined,linesnumbered]{algorithm2e}

\graphicspath{ {./images/} }

\title{Single agent robust deep reinforcement learning for bus fleet control}

\author{
  Yifan Zhang\thanks{Central South University, \href{mailto:erzhu419@gmail.com}{erzhu419@gmail.com}} \and
  Liang Zheng\thanks{Corresponding author. Central South University, \href{mailto:zhengliang@csu.edu.cn}{zhengliang@csu.edu.cn}} \and
  Qifan Zhang\thanks{Faculty of Information Technology, Monash University Malaysia, \href{mailto:zhangqifan0419@gmail.com}{zhangqifan0419@gmail.com}}
}
\date{}

\let\svthefootnote\thefootnote
\newcommand\blfootnotetext[1]{%
  \let\thefootnote\relax\footnote{#1}%
  \addtocounter{footnote}{-1}%
  \let\thefootnote\svthefootnote%
}

\let\svfootnotetext\footnotetext
\renewcommand\footnotetext[2][?]{%
  \if\relax#1\relax%
    \ifnum\value{footnote}=0\blfootnotetext{#2}\else\svfootnotetext{#2}\fi%
  \else%
    \if?#1\ifnum\value{footnote}=0\blfootnotetext{#2}\else\svfootnotetext{#2}\fi%
    \else\svfootnotetext[#1]{#2}\fi%
  \fi
}

\begin{document}
\maketitle

\begin{abstract}
	Bus bunching remains a critical challenge in urban transit systems, primarily driven by the stochastic nature of traffic conditions and passenger demand. Recently a popular method to address this issue is multi-agent reinforcement learning (MARL) applied in an idealized loop-line environment. However, such method generally suffers from high computational costs and sample inefficiency. Moreover, they often fail to capture the dynamics of realistic bus systems, which are typically governed by heterogeneous trip line and variable fleet sizes. In this study, we propose a robust single-agent reinforcement learning (RL) framework for bus holding control in a bidirectional timetabled bus line, explicitly designed to circumvent the data imbalance and convergence issues associated with MARL. Our key contribution focuses on transforming the inherently multi-agent problem into a single-agent formulation by explicitly encoding categorical identifiers---such as vehicle, station, and trip IDs alongside traditional numerical features (e.g., headway, occupancy and segment velocity) together as the state representation. This feature space augmentation enables the single-agent to operate effectively in a higher-dimensional space, analogous to projecting linearly inseparable inputs into a higher-dimensional space to achieve separability. Additionally, we introduce a structured "ridge-shaped" reward function that incentivizes the alignment with both uniform headways and scheduled departure intervals. Compared to the other three benchmark methods, our proposed RL algorithm achieves significantly more stable and higher rewards (\ensuremath{-430}k comparing with \ensuremath{-530}k) under stochastic passenger demands and inter-station travel time. These experimental results suggest that the proposed single-agent RL approach, informed by categorical identifiers in the state representation and realistic schedule-aware design in the ridge-shaped reward function, can effectively mitigate bus bunching in non-loop settings. This paradigm offers a robust and scalable alternative to those conventional MARL-based control frameworks, particularly in environments where agent-specific experience distributions are inherently imbalanced.
\end{abstract}


\section{Introduction}

Bus bunching remains a pervasive challenge in contemporary urban public transportation systems. This phenomenon occurs when two or more buses operating on the same route cluster together, thereby compromising service reliability and prolonging passenger waiting times. While often attributed simply to schedule deviations, empirical observations indicate a more intricate mechanism, driven by the dynamic interplay between traffic conditions and passenger demand \cite{bunching_wiki}. Notably, bunching tends to emerge not necessarily during peak rush hours, but often in the transitional intervals preceding them. During these periods, a sudden surge in passenger demand or a reduction in average road speed can significantly decelerate a leading bus, due to extended dwell times or congestion \cite{REZAZADA2024766}. Concurrently, the trailing bus, encountering fewer passengers or reduced congestion, begins to close the gap. If the trailing bus eventually overtakes the leader, it then inherits the heightened demand at subsequent stops, decelerating in turn. This self-reinforcing cycle where the new leader is persistently burdened by increased demand results in buses alternately decelerating and accelerating relative to one another, eventually converging into a stable cluster \cite{wu2017modelling}. Over time, such unstable headway dynamics deteriorate the overall temporal regularity of the service. This phenomenon underscores the critical importance of developing control strategies that can adapt in real-time not only to current headway but also to demand and speed conditions that exhibit high volatility. Traditional methods, as well as conventional reinforcement learning (RL)-based approaches, often fail to accommodate the asynchronous, event-driven nature of these interactions. 

Recent advancements in bus fleet coordination have increasingly leveraged multi-agent reinforcement learning (MARL) frameworks, enabling individual vehicles to develop decentralized policies based on local observations. While MARL has demonstrated success in idealized environments such as loop-line or circular routes where the fleet size and trip distribution remain static, its applicability to realistic, bidirectional, and timetable-governed networks is hindered by several fundamental limitations.

Theoretically, MARL-based approaches are highly susceptible to severe data imbalance across agents. This phenomenon is particularly evident in real-world operations where buses deployed exclusively during peak hours (e.g., the 13\textsuperscript{th} bus in a morning rush) accumulate substantially fewer experiences compared to those operating continuously, leading to unstable or suboptimal policy convergence. Furthermore, unlike continuous loop-line settings, bidirectional lines are inherently finite and bounded by terminal stops. This structural difference prevents agents from accumulating rewards over multiple loops, thereby impeding effective credit assignment and the propagation of long-term rewards across truncated episodes. This challenge is further exacerbated by the asynchronous, event-driven nature of transit systems. Unlike traditional MARL benchmarks like robotic soccer, which rely on synchronous decision-making, bus operations involve agents making decisions at irregular, temporally misaligned intervals\cite{wang2021reducing, wang2020dynamic}.

Our research addresses these gaps by introducing a robust single-agent deep reinforcement learning framework explicitly tailored to the complexities of realistic, timetable-based operations. By reformulating the inherently multi-agent task into a single-agent problem through the use of categorical identifiers (such as vehicle and station IDs), we circumvent the data imbalance and convergence issues typical of MARL. This approach enables a unified policy to generalize across a heterogeneous fleet, effectively learning a robust control law that accounts for the asynchronous interactions and finite trip structures inherent in bidirectional transit corridors.

In pursuit of this objective, this paper presents four key contributions. First, we have developed a realistic bidirectional bus simulation environment designed to bridge the gap between theoretical RL research and practical transit deployment. Unlike the commonly employed loop-line settings in previous studies, which assume fixed fleet sizes and uniform trip distribution among vehicles across the fleet, our environment operates under real-world constraints such as time-varying demand, asymmetric direction stops, stochastic traffic conditions, and dynamic fleet activation based on timetable triggers. Second, we propose a novel single-agent RL framework that reformulates the multi-agent bus control problem as a single-agent task. By explicitly embedding categorical features (e.g., vehicle ID, stop ID, and trip ID) and concatenating them with continuous features such as headway and segment speed, our model enables the agent to generalize across heterogeneous buses and temporal contexts. This design leverages the intuition behind high-dimensional feature mappings analogous to kernel methods in machine learning where linearly inseparable patterns become linearly separable once augmented with sufficient structural feature dimensions. Third, we design a structured reward function that replaces common exponential-based heuristics with a ``ridge-shaped'' topology, which simultaneously incentivizes uniform headways and adherence to the scheduled inter-vehicle departure intervals. This design draws upon principles from transit service planning literature \cite{ceder2016public}, emphasizing that the schedule itself is the core determinant of service quality. Fourth, we empirically demonstrate that soft actor critic (SAC), even in its standard form, provides inherent robustness in stochastic and dynamically disturbed environments, aligning with recent theoretical findings that connect maximum entropy RL with robustness guarantees \cite{Eysenbach2022MaxEnt}. Extensive experiments confirm that our SAC-based approach outperforms the MARL baseline multi-agent deep deterministic policy gradient (MADDPG) by achieving reduced variance and superior reward accumulation, particularly under peak-period stress scenarios.

The structure of this paper is organized as follows: Section 2 provides a review of related literature on bus bunching issues and reinforcement learning methodologies within public transportation systems, with a particular focus on bus operations. Section 3 outlines the problem formulation and simulation environment, incorporating the bus operation model and stochastic passenger demand. Section 4 details the proposed method, specifying the reinforcement learning framework and reward formulation. Section 5 presents the experimental results and performance comparisons. Finally, Section 6 concludes the paper and discusses directions for future research.

\section{Related work}

\subsection{Bus bunching mitigation strategies}
Bus bunching mitigation strategies have traditionally been categorized into station-based and inter-station-based control approaches~\cite{ceder2016public}. Station holding aims to regulate headways by delaying early-arriving buses~\cite{cats2012bus}, while inter-station control adjusts cruise speeds or leverages traffic signal priority to maintain spacing~\cite{bie2020dynamic}. Recent achievements have explored integrated, hybrid strategies that synthesize both control dimensions. For instance, \cite{Nie2024MultiStrategy} proposed a multi-strategy system leveraging deep reinforcement learning to unify stop-level holding, speed guidance, and signal adaptation. While such integration improves adaptability under complex conditions, it often introduces training instability, especially in high-dimensional or temporally asynchronous environments. Furthermore, a significant portion of the existing literature assumes closed-loop or loop-line configurations~\cite{cortes2010hybrid, cats2019frequency}, which oversimplifies realistic bidirectional and timetable-governed operations.

Our work addresses these limitations by developing a learning architecture compatible with flexible scheduling and dynamic fleet sizes. Traditional bus control strategies often fail to fully account for the inherent uncertainties in urban transit systems, such as fluctuating passenger demand and stochastic travel times. To address this, recent research has shifted toward robust decision-making frameworks. For instance, \cite{Zheng2024RobustNonlinearBusSpeedControl} developed a robust nonlinear decision mapping approach for online bus speed control under uncertainty, which was further extended by \cite{Liu2024RobustBus} into a bi-objective optimization framework that accounts for both implementation errors and traffic volatility. While these approaches demonstrate the efficacy of robust mapping in speed regulation, they often rely on static control laws. In contrast, our work leverages the adaptive nature of SAC to learn a robust station-based holding policy, explicitly tailored to handle the complex dynamics of bidirectional, timetabled networks under extreme stochastic conditions.

\subsection{Reinforcement learning approaches in transit systems}
Reinforcement learning, particularly MARL, has emerged as a prominent tool for optimizing transit operations under stochastic passenger demand and traffic fluctuations. Notably, \cite{wang2021reducing} and \cite{Menda2019EventDriven} introduced asynchronous MARL frameworks designed to handle event-driven control through macro-actions, thereby significantly reducing policy horizon complexity. These efforts have been extended in hierarchical MARL studies such as \cite{Yu2024HMARL}, where high-level agents coordinate holding and acceleration actions within a decentralized paradigm.

Despite these advancements, MARL approaches often encounter scalability and data imbalance issues, especially when vehicles operate sporadically or are governed by temporally sparse policies. This phenomenon, frequently observed during peak-only deployments or irregular schedules, leads to convergence instability and sample inefficiency~\cite{wang2021reducing}. While hierarchical and curriculum-based RL frameworks~\cite{Tang2024SmartCities, Yu2024HMARL} have been proposed to address these issues, they often require extensive domain-specific design and expert priors.

In contrast, our work leverages a single-agent RL architecture with explicit encoding of categorical identifiers (e.g., time period, direction), enabling uniform policy learning across all vehicles regardless of deployment frequency. This transformation effectively mitigates data imbalance and improves generalization without necessitating multiple cooperating agents.

\subsection{Single-agent soft actor-critic and robustness guarantees}
SAC has emerged as a state-of-the-art off-policy RL algorithm known for its entropy-maximizing objective, which balances exploration and exploitation \cite{haarnoja2018soft}. In environments subject to noise and temporal disturbances, SAC has demonstrated superior sample efficiency and stability compared to deterministic policy gradients or Q-learning variants \cite{Eysenbach2022MaxEnt}.

Recent theoretical achievements \cite{Eysenbach2022MaxEnt}  have established that SAC approximately solves a robust reinforcement learning objective of the form:
\begin{equation}
	\max_{\pi} \min_{r \in \mathcal{R}, p \in \mathcal{P}} \mathbb{E}_{\pi,p} \left[ \sum_t r(s_t, a_t) \right],
\end{equation}
where the inner minimization spans a bounded uncertainty set over reward functions $\mathcal{R}$ and transition kernels $\mathcal{P}$. This equivalence indicates that entropy-regularized RL policies inherently exhibit robustness to bounded adversarial perturbations in both reward and dynamics.

Our approach builds on SAC, but extends it by embedding categorical features for developing policies in a highly stochastic bidirectional bus environment. By introducing time-varying origin-destination (OD) flows and Gaussian-distributed travel time disturbances, we aim to enhance the robustness of the learned policy. Combined with a structured, ridge-shaped reward function that emphasizes headway regularity and schedule adherence, our method not only achieves robust convergence but also aligns operational decisions with service quality objectives.

\subsection{(robust) \revise{O}ptimization-based approaches}
Recent simulation-optimization based studies have extensively investigated bus scheduling and operational strategies, extending to electric bus fleet management and rescheduling. For instance, \cite{Tang2023TRD} optimized a single-line electric bus fleet services by using skip-stop as the action, resolving the trade-off between efficiency and passenger demand. Furthermore, \cite{Tang2023Transportmetrica} explored hybrid transit services with modular autonomous vehicles, introducing a flexible and demand responsive service concept. The robustness of schedule performance was also examined through vehicle-type selection and departure-time shifting in electric bus routes \cite{Tang2023IJST}. At the fleet level, \cite{Tang2021TRR} investigated replacement strategies for variability kinds of electric buses, particularly, the heterogeneity in vehicle characteristics. Additionally, multi-line methodologies have been adapted for single-line operations to enhance the stability of fleet headway and passenger service quality \cite{Tang2020IJST}. While these contributions provide valuable optimization frameworks, they largely rely on deterministic formulations, which may face challenges in transit environments characterized by high uncertainty, such as fluctuating demand and travel time variability. To bridge this gap, we propose a reinforcement learning-based approach capable of robustly adapting to dynamic demand and uncertain travel conditions.

Robust simulation-based optimization has also been explored in multiobjective settings. For example, \cite{Zheng2024AOR} proposed robust approaches for constrained multiobjective problems, while \cite{Zheng2022EJOR} investigated biobjective robust optimization in unconstrained settings. These works highlight the critical relevance of robust optimization techniques for complex stochastic environments.

\section{Problem formulation and simulation environment}

\subsection{Bus operation mode}
This study considers a general bidirectional scheduled bus corridor system. The route comprises a set of stations $\mathcal{S} = \{1, 2, \dots, N\}$, including two terminal stations and $N-2$ intermediate stops. The system operates within a fixed service window $[T_{start}, T_{end}]$.

The vehicle dispatching process adheres to a pre-defined timetable spanning a predefined service window $T_{ops}$. Buses are dispatched at a constant scheduled headway $H$ in both directions. Service from the upstream terminal commences at $t_{start}$, while the downstream terminal starts with an offset $\Delta t$ to stagger departures. This offset time $\Delta t$ ensures an initial separation between the two directions, facilitating the analysis of headway evolution by minimizing inter-directional interference.

At each scheduled dispatch time:
\begin{itemize}
	\item If an idle vehicle is available at the terminal, it is assigned to the trip.
	\item If no vehicle is available, a new bus is activated.
\end{itemize}
Each vehicle executes a trip from one terminal to the other. Upon completion a trip, the vehicle becomes idle and available for subsequent assignments. This mechanism inherently models dynamic fleet sizing and resource constraints.

\subsection{Passenger demand and traffic dynamics}
Passenger demand is characterized by Time-Dependent Origin-Destination (OD) flows. Let $\lambda_{mn}(t)$ denote the arrival rate of passengers traveling from stop $m$ to stop $n$ during time period $t$. Passenger arrivals are assumed to follow a Poisson process driven by these time-varying rates.

Road traffic conditions are modeled as stochastic variables. The travel time or speed between adjacent stops $m$ and $m+1$ follows a distribution $\mathcal{V}_{i,t}$ (e.g., Gaussian) defined by a time-varying mean $\mu_{v}(t)$ and variance $\sigma_{v}^2(t)$. This stochasticity captures real-world disturbances such as congestion and traffic signal delays.

\subsection{Illustrative example of bunching dynamics}
To illustrate the instability of open-loop bus operations, consider two consecutive vehicles, Bus $k$ and Bus $k+1$, scheduled with headway $H$. The \textbf{Bus Bunching} phenomenon emerges from a positive feedback loop:
\begin{itemize}
    \item \textbf{Initial Perturbation}: Suppose Bus $k$ is slightly delayed by a stochastic disturbance (e.g., traffic signal).
    \item \textbf{Passenger Accumulation}: The arrival interval at the next station extends ($h_k > H$), causing excess passengers to accumulate beyond expectation.
    \item \textbf{Dwell Time Expansion}: Increased boarding demand requires prolonged dwell time, further delaying Bus $k$ (the ``slow get slower'' effect).
    \item \textbf{Compression of Following Headway}: Conversely, Bus $k+1$ arrives closer to Bus $k$ ($h_{k+1} < H$), encounters fewer waiting passengers, and dwells for a shorter time (the ``fast get faster'' effect).
\end{itemize}
Consequently, the gap $h_{k+1}$ diminishes continuously until the two buses clump together. This inherent instability necessitates active control strategies to break the loop.

\subsection{Decision timing and state transition}
\begin{figure}[htbp]
	\centering
  \phantomsection 
	\includegraphics[trim=130 400 130 195, clip, width=0.85\textwidth]{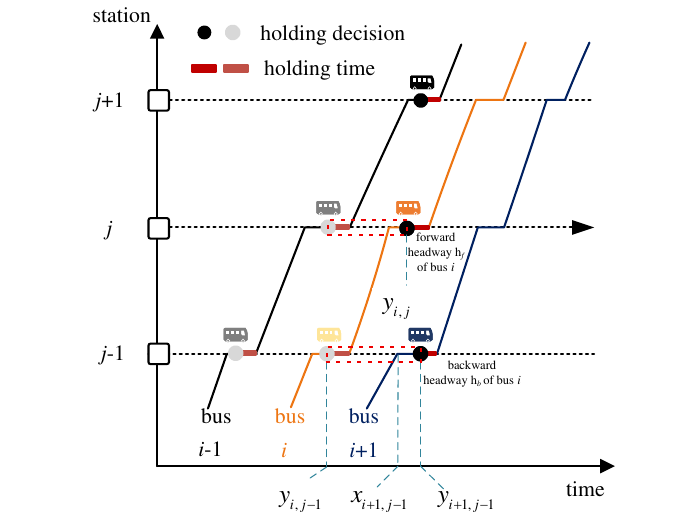}
	\caption{The timing of sequential decision and feedback.}
	\label{fig:decisiontiming}
\end{figure}

Unlike traditional methods that trigger decisions immediately upon bus arrival at a stop (denoted $x_{i+1,j-1}$ in Fig.~\ref{fig:decisiontiming}) \cite{cortes2010hybrid,Zheng2024RobustNonlinearBusSpeedControl}, our model strategically postpones decision-making to a post-service epoch after all passenger boarding and alighting activities have completed. This moment, denoted as $y_{i+1,j-1}$ in Fig.~\ref{fig:decisiontiming}, aligns more robustly with operational control opportunities in practice.

At $y_{i,j-1}$, the agent receives a state observation and generates a control action. The environment then executes the action (e.g., holding), and upon reaching the subsequent stop at $y_{i,j}$, the agent receives a reward based on the resulting system state (e.g., headway smoothness). This enables the RL agent to respond to the most current post-service information, leading to more stable and informed control decisions.

Figure~\ref{fig:decisiontiming} illustrates the holding decision process and headway computation using three sample buses traveling along a corridor. Each polyline represents the trajectory of an individual bus, where time progresses along the horizontal axis and station index increases along the vertical axis. The darkest bus symbol on each line denotes the most recently visited stop, while progressively lighter segments indicate historical stops previously visited. Holding durations are overlaid along each segment: \textcolor{red}{red} bars indicate the current holding time decision, and \textcolor{Bittersweet}{orange} bars denote previous holding actions.

The dash line \textcolor{red}{red} rectangle in the figure highlights a critical region used to compute both forward and backward headways. Specifically, the horizontal time difference between \( y_{i,j-1} \) and \( y_{i+1,j-1} \) quantifies the forward headway \( h_f \) of bus \( i \) and simultaneously the backward headway \( h_b \) of bus \( i+1 \). This quantity becomes observable only when \textbf{both} buses have completed dwell operations at the same station, ensuring that the headway reflects the realized temporal interval between two consecutive active vehicles.

\section{Methodology}
\subsection{Overall methodology framework}
\label{sec:methodology-framework}

\revise{Our control framework transforms the multi-agent bus holding problem into a single-agent reinforcement learning task. To address the heterogeneity of diverse vehicles and stops within a unified policy, we employ categorical embedding layers to map discrete identifiers (e.g., \texttt{bus\_id}, \texttt{stop\_id}) into dense latent representations. This allows a single SAC agent to control the entire fleet by generalizing across varied spatio-temporal contexts.}

\revise{The core control logic is built upon an event-driven variant of the soft actor-critic (SAC) algorithm. Unlike standard step-synchronous RL where decisions occur at fixed time intervals, our agents' actions and rewards are triggered asynchronously by \emph{bus-arrival events} at stations.}

\revise{As illustrated in Algorithm~\ref{alg:sac-bus}, we maintain two dictionaries indexed by \texttt{bus\_id}: \texttt{action\_dict} (to store the state-action pair $(s_t,a_t)$ at the moment of decision) and \texttt{state\_dict} (to track the latest observable state). A complete transition tuple $(s_t, a_t, r_{t+1}, s_{t+1})$ is only committed to the replay buffer when the \emph{same} vehicle reaches its subsequent stop, ensuring the reward reflects the realized headway regularity.}

\begin{algorithm}[H]
\DontPrintSemicolon
\SetAlgoLined
\caption{Event-Driven Categorical Feature SAC}
\label{alg:sac-bus}

\KwIn{Environment $\mathcal{E}$, target headway $\tau$, penalty threshold $\delta$, maximum holding duration $H_{max}$, replay buffer $\mathcal{D}$, batch size $B$, target update rate $\rho$, temperature $\alpha$.}
\KwOut{Policy $\pi_\theta(a\mid s)$ and critics $Q_{\phi_1},Q_{\phi_2}$.}

\BlankLine
\textbf{Initialization:} Randomly initialize network weights $\theta,\phi_1,\phi_2$ and target weights $\bar\phi_1,\bar\phi_2$; $\mathcal{D} \leftarrow \emptyset$; \texttt{state\_dict} $\leftarrow \{\}$, \texttt{action\_dict} $\leftarrow \{\}$.\;
\textbf{Environment Reset:} Initialize $\mathcal{E}$ based on the predefined timetable; obtain initial states for active buses; update \texttt{state\_dict}.\;

\While{\text{simulation not done}}{
  \tcp{(A) Event progression: advance to next arrival}
  $(\texttt{bus\_id}, s^{\text{arr}}_{t+1}) \leftarrow \mathcal{E}.\textsc{StepToNextArrival}()$\;
  \texttt{state\_dict[bus\_id]} $\leftarrow s^{\text{arr}}_{t+1}$\;

  \tcp{(B) Experience Assembly (Latching Mechanism)}
  \If{\texttt{bus\_id} $\in$ \texttt{action\_dict}}{
      $(s_t,a_t) \leftarrow$ \texttt{action\_dict[bus\_id]}\;
      $r_{t+1} \leftarrow \textsc{Reward}(s_t, a_t, s^{\text{arr}}_{t+1}; \tau, \delta)$ \tcp*[r]{General ridge-shaped reward}
      $d \leftarrow \textsc{IsTerminal}(s^{\text{arr}}_{t+1})$\;
      $\mathcal{D}.\textsc{Push}(s_t, a_t, r_{t+1}, s^{\text{arr}}_{t+1}, d)$\;
      \texttt{delete action\_dict[bus\_id]}\;
  }

  \tcp{(C) Decision Making (Action $a \in [0, H_{max}]$)}
  $s_{t+1} \leftarrow$ \texttt{state\_dict[bus\_id]}\;
  $a_{t+1} \sim \pi_\theta(\cdot \mid s_{t+1})$ mapped to $[0, H_{max}]$\;
  \texttt{action\_dict[bus\_id]} $\leftarrow (s_{t+1}, a_{t+1})$\;
  $\mathcal{E}.\textsc{ApplyHolding}(\texttt{bus\_id}, a_{t+1})$\;

  \tcp{(D) Model Updates}
  \If{$|\mathcal{D}| \ge B$}{
      Sample batch from $\mathcal{D}$ and update $\phi_i, \theta, \alpha$ using the SAC objective (Eqs. 4--7).;
      Update target networks: $\bar\phi_i \leftarrow \rho \phi_i + (1-\rho)\bar\phi_i$.\;
  }
}
\end{algorithm}

\revise{
\noindent \textbf{Remarks.} 
(i) The asynchronous tuple assembly prevents temporal leakage by pairing actions with their actual realized outcomes. 
(ii) The dictionaries function as a per-bus latch, accommodating a varying number of active vehicles without requiring a multi-agent formulation. 
(iii) Parameters such as $\tau$ and $H_{max}$ are environment-specific and are detailed in Section~\ref{sec:experiment}.
}
\subsection{Components of reinforcement learning}

Rather than assigning an individual policy to each vehicle, the system employs a single reinforcement learning agent that observes the operational context of each bus at decision points and outputs control actions accordingly. Unlike conventional time-stepped RL settings, decisions here are triggered by discrete events specifically, upon completion of passenger boarding and alighting at a stop. Let \( x_{i,j} \) denote the time when bus \( i \) arrives at stop \( j \), and \( y_{i,j} \) be the moment when all passenger activities are completed. At time \( y_{i,j} \), the agent receives a state observation \( s_{i,j} \), obtains a reward \( r_{i,j} \), and selects a continuous holding action \( a_{i,j} \in [0, H_{max}] \). This action determines how many additional seconds the vehicle will hold at the stop after natural dwelling, with the goal of improving headway regularity. The subsequent state \( s'_{i,j} \) is only observed when bus \( i \) completes boarding and alighting at stop \( i \), i.e., at \( y_{i,j+1} \). The entire process is illustrated in Fig.~\ref{fig:decisiontiming}.

\paragraph{State Representation.}
The state vector \( s_{i,j} \in \mathbb{R}^{d} \) includes both categorical and numerical features. The categorical component consists of:
\begin{equation}
\texttt{[bus\_id, station\_id, time\_period, direction]}
\end{equation}
where \texttt{time\_period} indicates the current hour of operation derived from total elapsed time $t$. These are encoded through learnable embedding layers to accommodate nonlinearity and improve generalization. The numerical features are:
\begin{equation}
\texttt{[forward headway, backward headway, current segment speed]}
\end{equation}
which directly reflect local traffic and service level explicitly. This hybrid structure enables the policy network to incorporate heterogeneous information sources and handle data imbalance arising from irregular trip activation. This address a common challenge in real-world systems where specific vehicles operate solely during peak periods, resulting in skewed experience distributions across agents. By embedding categorical variables such as bus and trip identifiers, the single-agent framework generalizes across heterogeneous spatio-temporal contexts without requiring separate parameterizations for each vehicle instance.

\paragraph{Action Definition.}
While inter-station-based speed control could theoretically provide finer granularity in maintaining regular headways, we deliberately adopt station-based holding control as our primary action space. This choice is driven by the following practical constraints observed in real-world bus operations:

\begin{itemize}
    \item \textbf{Safety and road conditions}: Dynamic speed adjustment is constrained by urban traffic rules, varying road conditions, and safety requirements. Buses operating in mixed traffic environments are restricted from arbitrarily accelerating or decelerating.
    
    \item \textbf{Vehicle inertia}: Acceleration or deceleration behavior is affected by bus occupancy levels. Heavily loaded buses exhibit higher inertia, limiting their responsiveness to speed commands.
    
    \item \textbf{Action uncertainty}: In realistic scenarios, instructions to modify speed are characterized by interpretation and execution by human drivers, introducing uncertainty in action realization. This makes direct speed control less reliable than station-based holding.
    
    \item \textbf{Operational preference}: Bus companies are more inclined to adopt stop-level holding strategies since these can be clearly communicated to drivers and executed unambiguously. Many transit agencies already implement static holding as part of daily operations.
\end{itemize}

Therefore, the action in our reinforcement learning formulation corresponds to holding time at the current station. This is not only operationally feasible and enforceable but also aligns with industry best practices for mitigating bus bunching while minimizing disruption to passenger experience.

So we define the action \( a_{i,j} \) as a scalar value sampled from a bounded continuous space \( [0, H_{max}] \), where \( H_{max} \) is a preset upper limit (detailed in Section~\ref{sec:experiment}). It corresponds to the additional dwell time (holding) a bus will incur beyond the necessary passenger exchange duration. This continuous formulation contrasts with discrete or macro-action settings in prior studies, allowing finer control granularity \cite{cortes2010hybrid, cats2012bus}.

\paragraph{Reward Function Design}
\begin{figure}[htbp]
    \centering
    \includegraphics[trim= 0 0 0 10,width=0.7\textwidth]{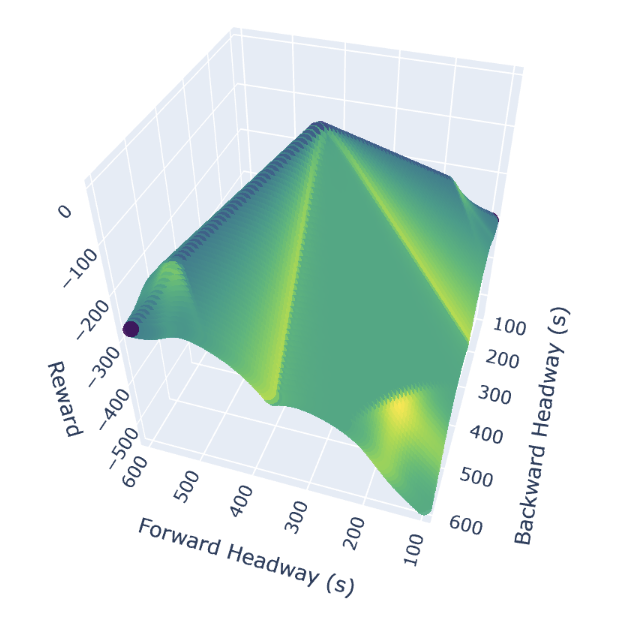}
    \caption{Visualization of the reward surface}
    \label{fig:rewardlandscape}
\end{figure}

The reward is designed to promote both symmetry in headways and schedule adherence. At each stop \( n \), once a bus \( i \) completes its boarding process, it receives a scalar reward \( r_{i,n} \) based on its \( h_f \) and \( h_b \). The function consists of three components:

\begin{itemize}
    \item \textbf{Schedule alignment}: A soft penalty proportional to the deviation from the nominal target headway of $\tau$.
    \item \textbf{Headway symmetry}: A similarity bonus inversely proportional to the absolute difference between \( h_f \) and \( h_b \), encouraging platoon balance.
    \item \textbf{Robust penalization}: Additional penalty is applied when either headway deviates from the target by more than $\delta$.
\end{itemize}

Let
\begin{equation}
	R(h_f, h_b) = \omega(h_f, h_b) \cdot \phi(h_f) + (1 - \omega(h_f, h_b)) \cdot \phi(h_b) - 0.5 \cdot |h_f - h_b| - 20 \cdot \mathbb{I}_{|h_f - \tau| > \delta \text{ or } |h_b - \tau| > \delta}
\end{equation}
where:
\begin{equation}
\phi(h) = -|h - \tau|, \quad \omega(h_f, h_b) = \frac{|h_f - \tau|}{|h_f - \tau| + |h_b - \tau| + \epsilon}
\end{equation}

Here \( \phi(\cdot) \) rewards headways close to $\tau$, while \( \omega \) dynamically weighs the forward and backward components. The final penalty term accounts for extreme outliers.

As illustrated in Fig.~\ref{fig:rewardlandscape}, the reward surface exhibits a prominent ridge along the line \( h_f = h_b = \tau \), where vehicle spacing is both balanced and schedule-aligned. Conversely, deviations along the direction \( h_f \ne h_b \) produce sharp declines in reward due to penalized asymmetry. This geometry ensures that the agent is incentivized not only to maintain consistent spacing, but also to adaptively adjust dwell times based on real-time headway context.
\revise{Although the reward surface is clearly ridge-shaped, a static plot may not fully convey its three-dimensional geometry. For better intuition, an interactive and rotatable visualization of the reward function is available as the \texttt{smoothed\_reward\_surface.html} file in our public repository at \url{https://github.com/erzhu419/Categorical-Feature-sac-in-bus-simulation}, where all experimental codes and assets used in this study are also provided.}

Although many prior studies advocate inter-station cruise speed control \cite{Daganzo2011cooperation,Nie2024MultiStrategy}, we explicitly choose station-based holding in this work due to three pragmatic considerations rooted in real-world operations:

\begin{enumerate}
\item Limited action executability: Speed control inherently requires the driver or vehicle control unit to continuously adjust velocity profiles in response to subtle policy outputs. In realistic bus operations, this is often impractical: driver response is delayed, bounded by road conditions, and strongly affected by onboard load-induced inertia. Consequently, the actual control action ( e.g., reducing speed or accelerating by 10\%) may not be reliably executed, introducing action uncertainty into the control loop.

\item Safety and regulatory constraints: Rapid speed modulation to maintain headway (especially deceleration) often conflicts with traffic safety rules and leads to passenger discomfort, which transit agencies are keen to avoid. This makes speed control operationally and contractually infeasible in many public transit settings.

\item Holding as the de facto industry norm: In practice, holding buses at stations (especially terminal or major stops) is a widely accepted operational strategy, as it is easier to communicate, enforce, and integrate into static schedules. Agencies also prefer holding because it enables pre-emptive coordination (e.g., scheduled rest periods or crew shifts) without increasing perceived risk.
\end{enumerate}

Therefore, we model bus control as discrete holding actions only, to better reflect feasible and enforceable policy deployment in real-world bidirectional corridors. While inter-station control remains a promising theoretical extension, its robustness under human-in-the-loop and partially controllable systems deserves separate investigation.

\paragraph{Transition and Experience Buffer.}
Due to the asynchronous nature of events, transitions \((s, a, r, s')\) cannot be immediately constructed. Instead, each agent maintains a temporary buffer indexed by its identifier. When a bus completes its stop interaction at \( y_{i,j} \), it stores \((s, a, r)\). Once the next state \( s' \) is available at \( y_{i,j+1} \), the full transition tuple \((s, a, r, s')\) is assembled and inserted into the global replay buffer. This design accommodates interleaved transitions from multiple agents and enables stable training in partially observable and event-driven environments.

\subsection{Feature representation and embedding network}

To enable a unified policy to generalize across different vehicles, stops, and temporal contexts, we explicitly incorporate four categorical variables into the state space:
\begin{equation}
\texttt{[bus\_id, stop\_id, direction, time\_period]}
\end{equation}
These features do not carry intrinsic numerical meaning but are critical for distinguishing instances with context-specific dynamics. Rather than employing one-hot encoding which would lead to high-dimensional sparse inputs, we adopt an embedding-based approach.

Each categorical variable is mapped to a dense, learnable vector through an embedding layer. Let \( c_j \) denote the value of the \( j \)-th categorical feature, and let \( E_j \in \mathbb{R}^{n_j \times d_j} \) be the corresponding embedding matrix, where \( n_j \) is the number of unique categories and \( d_j \) is the embedding dimension (typically \( d_j = \min(50, n_j/2) \)). Then the embedded representation is:
\begin{equation}
\mathbf{e}_j = E_j[c_j]
\end{equation}

All categorical embeddings are then concatenated:
\begin{equation}
\mathbf{e} = \texttt{concat}(\mathbf{e}_1, \mathbf{e}_2, \mathbf{e}_3, \mathbf{e}_4)
\end{equation}

This dense vector \( \mathbf{e} \) is further concatenated with the numerical part of the state vector:
\begin{equation}
\textit{s} = \texttt{concat}(\mathbf{e}, \textit{h}_f, \textit{h}_b, \mathbf{v}_{\text{segment}})
\end{equation}

where \( \mathbf{v}_{\text{segment}} \) is the current route segment speed. We also investigated including the speeds of all route segments as input features, but observed negligible improvement in performance.

\begin{figure}[htbp]
    \centering
    \includegraphics[trim = 60 420 200 100, clip, width=0.8\textwidth]{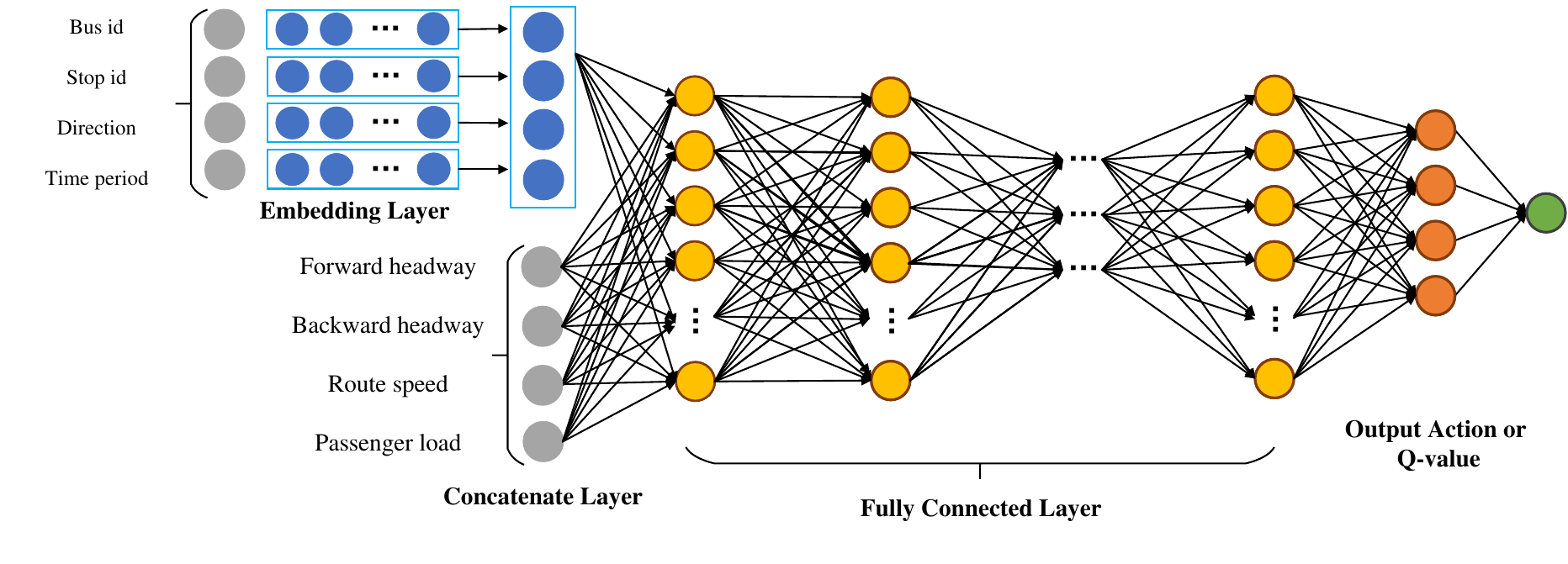}
    \caption{Network architecture}
    \label{fig:embedding-arch}
\end{figure}

The final input vector \( \textit{s} \in \mathbb{R}^d \) is fed into both the Q-networks and policy network in the SAC framework. Fig.~\ref{fig:embedding-arch} illustrates this process.

This architecture not only reduces the number of required agents thereby indirectly reducing the dimensionality of the learning problem, but also enables the network to capture semantic similarity between different context entities. For instance, two buses operating during the same peak hour or at adjacent stops can share latent representations via shared embedding weights, improving generalization and training stability. The use of entity embeddings is particularly important in our single-agent setting, where data distribution is inherently imbalanced across different vehicle instances.

\subsection{Soft actor-critic framework}

SAC is an off-policy deep reinforcement learning algorithm that integrates the maximum entropy principle into the actor-critic paradigm \cite{haarnoja2018soft, haarnoja2018applications}. Unlike traditional actor-critic algorithms that focus purely on maximizing cumulative reward, SAC augments the objective with an entropy regularization term. This encourages the learned policy to remain stochastic during training, which enhances exploration and naturally promotes more robust behavior in environments characterized by noise or partial observability.

Formally, sac optimizes the following objective:
\begin{equation}
J(\pi) = \sum_t \mathbb{E}_{(s_t, a_t) \sim \rho_\pi} \left[ r(s_t, a_t) + \alpha \mathcal{H}(\pi(\cdot | s_t)) \right] \tag{10}
\end{equation}
where \( \mathcal{H}(\pi(\cdot | s_t)) = -\mathbb{E}_{a_t \sim \pi(\cdot | s_t)} [\log \pi(a_t | s_t)] \) is the policy entropy, and \( \alpha \) is a temperature parameter balancing reward and entropy. In high variance stochastic environment setting, this structure is particularly useful for ensuring the policy does not collapse to greedy behavior too early during training, which is crucial given the asynchronous, unevenly distributed bus experience data.

To evaluate and update the policy, SAC relies on a soft version of the Bellman backup operator, which iteratively estimates the soft Q-function under the current policy:
\begin{equation}
Q^\pi(s_t, a_t) = r(s_t, a_t) + \gamma \mathbb{E}_{s_{t+1} \sim p} \left[ V^\pi(s_{t+1}) \right] \tag{11}
\end{equation}
\begin{equation}
V^\pi(s_t) = \mathbb{E}_{a_t \sim \pi} \left[ Q^\pi(s_t, a_t) - \alpha \log \pi(a_t | s_t) \right] \tag{12}
\end{equation}
Here, \( Q^\pi \) is the soft state-action value function, and \( V^\pi \) is the soft state value function implicitly defined by entropy adjustment. The role of the Bellman backup is to converge toward the true Q-values under policy \( \pi \), forming the core target signal for critic learning.

The critic networks are trained by minimizing the following squared soft Bellman residual:
\begin{equation}
J_Q(\theta_i) = \mathbb{E}_{(s_t, a_t, r_t, s_{t+1}) \sim \mathcal{D}} \left[ \left( Q_{\theta_i}(s_t, a_t) - y_t \right)^2 \right], \quad \text{where} \quad y_t = r_t + \gamma V^{\pi}_{\bar{\theta}}(s_{t+1}) \tag{13}
\end{equation}
In our implementation, two Q-networks \( Q_{\theta_1} \) and \( Q_{\theta_2} \) are trained simultaneously to reduce overestimation bias, and the target network value \( V^\pi \) is computed using the smaller of the two Q-values and a slowly updated target network \( \bar{\theta} \). This stabilizes critic convergence under highly variable headway dynamics and OD noise.

The policy \( \pi_\phi(a|s) \) is modeled as a squashed Gaussian distribution whose samples are passed through a \( \tanh \) nonlinearity to enforce bounded actions. This is critical in our domain, where the action represents additional holding time and must lie within a predefined range.

Policy learning is performed by minimizing the expected KL divergence between the policy and a softmax over the Q-function, originally expressed as:
\begin{equation}
	\pi^* = \arg\min_{\pi_\phi} D_{\mathrm{KL}}\left( \pi_\phi(\cdot|s) \;||\; \frac{\exp\left( Q_\theta(s, \cdot) \right)}{Z_\theta(s)} \right) \tag{14}
\end{equation}
Here, the normalization constant \( Z_\theta(s) = \int_A \exp(Q(s,a)/\alpha) \, da \) ensures that the exponentiated Q-function defines a proper probability density over the action space. Notably, \( Z_\theta(s) \) is independent of the current policy \( \pi \), and thus does not affect gradient-based updates.

By expanding the KL divergence and taking gradients w.r.t. \( \pi \), this yields the policy loss:
\begin{equation}
	J_\pi(\phi) = \mathbb{E}_{s_t \sim \mathcal{D}} \left[ \mathbb{E}_{a_t \sim \pi_\phi} \left[ \alpha \log \pi_\phi(a_t | s_t) - Q_{\theta}(s_t, a_t) \right] \right] \tag{15}
\end{equation}
which is the standard sac actor loss minimized during training. To avoid manual tuning of the temperature parameter \( \alpha \), SAC treats it as a learnable dual variable and updates it via dual gradient descent to enforce a target entropy. Specifically, \( \alpha \) is optimized by minimizing:
\begin{equation}
J(\alpha) = \mathbb{E}_{a_t \sim \pi} \left[ -\alpha \log \pi(a_t | s_t) - \alpha \bar{\mathcal{H}} \right] \tag{16}
\end{equation}
where \( \bar{\mathcal{H}} \) denotes the desired entropy level. This mechanism ensures that the policy maintains high stochasticity in the early stages of training and gradually becomes more deterministic as convergence is approached.
The inner term encourages the policy to choose actions with high Q-values while maintaining sufficient entropy.

SAC is well-suited to our asynchronous, event-driven dispatching environment. It allows for off-policy updates using irregular, vehicle-dependent transitions and gracefully handles nonstationary feedback signals. In contrast to MARL methods that require dense and balanced interactions across agents, our SAC implementation can learn a single policy that generalizes over time, space, and bus identity.

\subsection{Equivalence of SAC and robust reinforcement learning}

The primary challenge in our bus holding control problem stems from dynamic uncertainty: travel times between stops exhibit stochastic variability throughout the day due to congestion and irregular passenger demand. This manifests primarily as uncertainty in the transition dynamics rather than the reward function itself. From a reinforcement learning perspective, this is fundamentally a \textit{dynamically robust} decision problem where the policy must perform well even under small, adversarial perturbations to the dynamics model.

Directly addressing dynamic robust RL problems is often intractable, especially in continuous control settings. However, Theorem 4.2 in \cite{Eysenbach2022MaxEnt} offers critical insight: optimizing a maximum entropy RL objective with a modified reward function \( \bar{r}(s_t, a_t, s_{t+1}) \triangleq \frac{1}{T} \log r(s_t, a_t) + \mathcal{H}[s_{t+1} \mid s_t, a_t]\) under the unperturbed dynamics \( p \) establishes a lower bound on the robust objective under the true (perturbed) dynamics \( \tilde{p} \). Specifically, instead of optimizing:

\begin{equation}
\max_{\pi} \min_{\tilde{p} \in \Delta_P} \mathbb{E}_{\pi, \tilde{p}} \left[ \sum_t r(s_t, a_t) \right], \tag{17}
\end{equation}

we optimize:

\begin{equation}
\mathbb{E}_{\pi, p} \left[ \sum_t \bar{r}(s_t, a_t) \right], \tag{18}
\end{equation}

which serves as a tight lower bound on the robust formulation.

As the agent accumulates sufficient trajectories such that the sample mean of \( r(s,a) \) stabilizes to a constant across time the Jensen gaps introduced in the theoretical derivation (see Eq. 9 in \cite{Eysenbach2022MaxEnt}) become small, and the lower bound becomes increasingly tight. This behavior is analogous to Bayesian optimization, where high posterior variance early in training warrants broad exploration, but posterior uncertainty shrinks as more samples are acquired. Similarly, the stochastic policy induced by SAC begins with high entropy (large \( \alpha \)) to hedge against uncertainty in reward and dynamics, and gradually converges to a more deterministic policy as uncertainty reduces.

Admittedly, this analogy is imperfect Bayesian posterior variance captures epistemic uncertainty, while real-world bus dynamics involve irreducible aleatoric uncertainty (e.g., signal timing fluctuations or passenger boarding variability). Nonetheless, the mechanism aligns well: early-stage high entropy enables SAC to explore and tolerate dynamics variability, while later-stage entropy decay naturally aligns the policy with the most likely transition behavior.

By leveraging SAC, we implicitly optimize a provably grounded surrogate for the dynamic robust control problem we care about without explicitly modeling worst-case perturbations. This makes SAC particularly well-suited to our setting, and empirically, it demonstrates greater stability than standard MARL baselines under the same simulation conditions.

\section{Experimental setup and results}
\label{sec:experiment}

\subsection{Experimental setup and hyperparameters}

To evaluate the performance and robustness of our proposed holding control method under realistic transit conditions, we develop a custom discrete-event simulation environment that models bidirectional bus operations with non-stationary(time-varying) demand and traffic uncertainty. The experimental setup is structured as follows.\revise{(You can find our source code at \url{https://github.com/erzhu419/Categorical-Feature-sac-in-bus-simulation.git})}

\subsubsection{Simulation environment and data sources}

The simulator is implemented in Python 3.8 and is designed to simulate dynamic bus fleet operations along a corridor with two terminal stations. Each simulation episode begins at 6:00 AM and runs until all scheduled trips in the daily timetable are completed.

Passenger demand and traffic conditions are synthetically generated following the modeling framework proposed in Zheng et al.~\cite{Zheng2024RobustNonlinearBusSpeedControl}. Specifically:

\begin{itemize}
    \item \textbf{Passenger demand} is specified by an hourly OD matrix stored in \texttt{passenger\_OD.xlsx}. For each pair of stations and each time period, the matrix specifies the expected number of passengers boarding per hour. During simulation, passenger arrivals at each stop are sampled from a Poisson distribution with a rate derived from the OD entry divided by 3600, generating fine-grained second-by-second demand.
    
    \item \textbf{Traffic speed} is provided in \texttt{route\_news.xlsx}, which contains the average speed (in m/s) for each inter-stop segment, varying by hour. Actual travel speed during simulation is sampled from a Gaussian distribution with the recorded mean and a fixed standard deviation of $1.5$, reflecting random perturbations in traffic conditions. This setup closely mirrors the uncertainty modeling approach in~\cite{Zheng2024RobustNonlinearBusSpeedControl}.
    
    \item \textbf{Route structure} is described in \texttt{stop\_news.xlsx}, which defines 22 stops including two terminals and 20 intermediate stops. Each vehicle operates in a single direction per trip, either from terminal\_down to terminal\_up (``up'' direction) or the reverse (``down'' direction).
    
    \item \textbf{Timetable-driven dispatching} is defined in \texttt{time\_table.xlsx}. The operation starts at 6:00 AM. Vehicles are dispatched every 360 seconds (6 minutes) from both terminals. An offset of 180 seconds is applied to the downstream direction to stagger departures (e.g., Upstream starts at 6:00:00, Downstream at 6:03:00). Once a vehicle completes a trip and returns to its terminal, it may either rest or be reassigned to a subsequent trip contingent on schedule availability.
\end{itemize}

This bidirectional, schedule-triggered structure extends the unidirectional simulation approach in~\cite{Zheng2024RobustNonlinearBusSpeedControl}, and enables the modeling of peak vs.\ off-peak asymmetry, heterogeneous headway patterns, and dynamic fleet scaling.

\textbf{Specific Case Study Parameters.} In the experiments presented in this section, we instantiate the general model with the following reliable real-world settings:
\begin{itemize}
    \item Service window $T_{ops} = 13$ hours (06:00 to 19:00);
    \item Scheduled headway $H = 360$ seconds;
    \item Dispatch offset $\Delta t = 180$ seconds;
    \item Number of stops $N = 22$.
\end{itemize}

\subsubsection{Network architectures and hyperparameters}

All models are implemented using PyTorch. The architectural and training configurations are:

\paragraph{Embedding-SAC\revise{, DDPG and TD3(single-agent RL)}}
\begin{itemize}
    \item \textbf{Embedding layers}: for each categorical variable (bus ID, station ID, time period, direction), a learnable embedding is constructed with dimension set as $\min(50, \lfloor N_i / 2 \rfloor)$, where $N_i$ is the number of unique values for category $i$;
    \item \textbf{Actor and critic networks}: both are 4-layer multilayer perceptrons (MLPs) with hidden sizes of [32, 32, 32], followed by task-specific output layers (mean/log-std for policy; scalar value for Q-function);
    \item \textbf{Activation function}: ReLU;
    \item \textbf{Optimizer}: Adam;
    \item \textbf{Learning rate}: $1\times10^{-5}$;
    \item \textbf{Batch size}: 2048;
    \item \textbf{Target smoothing coefficient}: 0.005.
\end{itemize}

\paragraph{MADDPG (PS/NPS)} 
\begin{itemize}
    \item \textbf{Actor and critic networks}: 3-layer MLPs with layer sizes \([64, 64, \text{output\_dim}]\) (where \(\text{output\_dim}\) is determined by the action or Q-value dimension);
    \item \textbf{Activation function}: ReLU;
    \item \textbf{Optimizer}: Adam; 
    \item \textbf{Learning rate}: $1\times10^{-5}$ for both actor and critic; 
    \item \textbf{Batch size}: 128; 
    \item \textbf{Target smoothing coefficient}: 0.01 (Polyak averaging); 
    \item \textbf{Parameter sharing}: In PS, all agents share a single actor and critic network, with agent IDs encoded as part of the input; in NPS, each agent trains its own independent actor and critic networks. 
    \item \textbf{Exploration noise}: Gaussian noise with standard deviation 0.2, clipped to the action bounds; 
    \item \textbf{Replay buffer size}: $1\times10^6$ transitions per agent. 
\end{itemize}

\paragraph{Batch Size Selection.}
The batch size settings for MADDPG and \revise{single-agent RL} differ significantly because the agents in MARL algorithms often share the data in replay buffer they collected from their individual experiences. Conversely in \revise{single-agent RL}, there is only one agent and one buffer. Every experience is stored in this single buffer, and the batch size is scaled according to the total number of experiences available. To maintain the same number of available transition tuples for every single-agent, which prevents any single-agent from being overwhelmed by the experiences of others, the batch size varies substantially between the single-agent RL and the MARL algorithms. 

In contrast, \revise{single-agent RL} employs a shared replay buffer across all buses, which allows it to sample from a larger and more diverse dataset. This design enables \revise{single-agent RL} to leverage experiences from other buses in the same fleet, improving generalization and stability. However, the most relevant data for \revise{single-agent RL} still comes from experiences associated with the same bus ID and station ID as the current decision point. Given that the fleet size is approximately 20 (varying slightly with traffic conditions), dividing the \revise{single-agent RL} batch size by the fleet size provides an effective state-action batch size comparable to MADDPG's setting. For example, with a \revise{single-agent RL} batch size of 2048, the effective per-bus batch size is approximately \( \frac{2048}{20} \approx 128 \), aligning with MADDPG's batch size.

This approach balances the benefits of global experience sharing with the need for local relevance, ensuring that \revise{single-agent RL} maintains both robustness and specificity in its policy updates.

\subsection{Performance comparison}

\begin{figure}[htbp]
  \centering
  \del{\mbox{\includegraphics[width=0.95\textwidth]{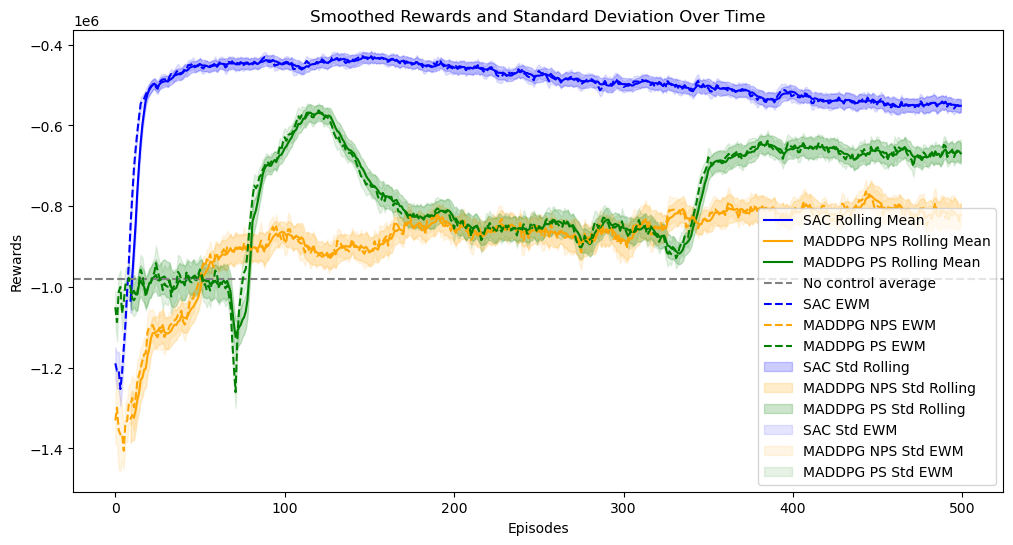}}}
  \revise{\includegraphics[width=0.95\textwidth]{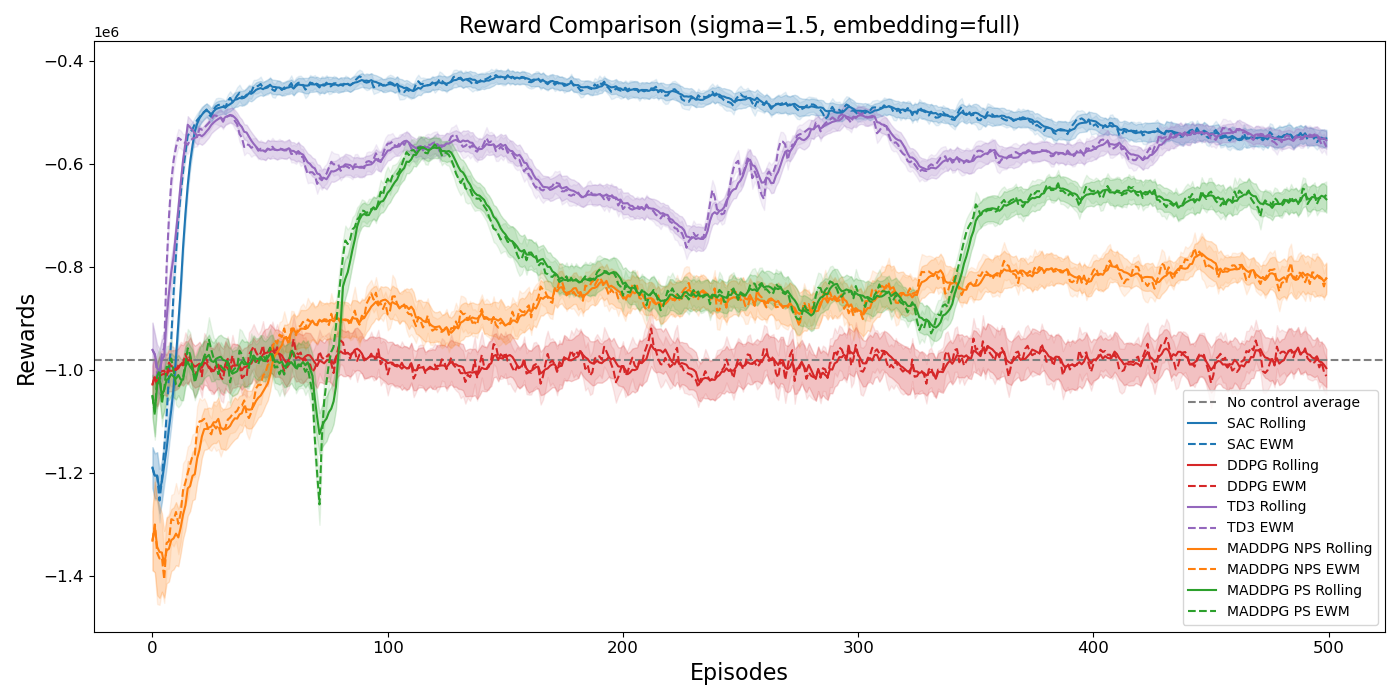}}
  \caption{Smoothed training rewards and standard deviations of \revise{single-agent RL} and MADDPG variants. Solid lines: 10-episode rolling means. Dashed lines: exponentially weighted moving averages (EWM, $\alpha=0.3$). Shaded areas: $\pm 1\revise{.5}$ standard deviation over 15 evaluation rollouts. Gray dashed line: average reward under uncontrolled dispatch.}
  \label{fig:performance_comparison}
\end{figure}

Figure~\ref{fig:performance_comparison} presents the training reward trends of the \revise{single-agent RL} method versus two variants of MADDPG: one with parameter sharing (MADDPG-PS) and one without (MADDPG-NPS). To better reveal convergence dynamics and stability, we display both 10-episode rolling means (solid lines) and exponentially weighted moving averages (EWM, dashed lines), along with shaded areas indicating the corresponding standard deviations. A horizontal dashed gray line is included to represent the average reward of the uncontrolled case, serving as a performance baseline for comparison.

{\del{Several key observations can be made from the results:}}

\begin{table*}[htbp]
\centering
\caption{\revise{Cross-variance evaluation of SAC, TD3, and DDPG under different embedding configurations. Values represent mean episode rewards (higher is better); best results in each column are \textbf{bolded}.}}
\label{tab:crosssigma}
\revise{%
\scalebox{1.0}{%
\begin{tabular}{llccc}
\toprule
\multicolumn{5}{c}{\textbf{\(\sigma_{\text{test}}=1.0\)}} \\
\midrule
Algorithm & Embedding & \(\sigma_{\text{train}}=1.0\) & \(\sigma_{\text{train}}=1.5\) & \(\sigma_{\text{train}}=2.0\) \\
\midrule
SAC & Full & -369.1K $\pm$ 9.6K & -347.4K $\pm$ 13.2K & \textbf{-322.8K $\pm$ 9.1K} \\
SAC & One-hot & -362.3K $\pm$ 9.7K & \textbf{-341.5K $\pm$ 10.5K} & -350.9K $\pm$ 8.5K \\
SAC & None & \textbf{-343.5K $\pm$ 12.7K} & -347.5K $\pm$ 13.4K & -366.2K $\pm$ 10.7K \\
DDPG & Full & -714.5K $\pm$ 29.6K & -698.7K $\pm$ 42.1K & -712.6K $\pm$ 21.2K \\
DDPG & One-hot & -708.1K $\pm$ 31.9K & -698.6K $\pm$ 32.7K & -690.9K $\pm$ 31.8K \\
DDPG & None & -698.9K $\pm$ 30.7K & -721.2K $\pm$ 19.1K & -698.3K $\pm$ 30.0K \\
TD3 & Full & -714.0K $\pm$ 33.6K & -492.1K $\pm$ 17.9K & -701.6K $\pm$ 28.6K \\
TD3 & One-hot & -706.9K $\pm$ 24.2K & -703.9K $\pm$ 32.5K & -709.6K $\pm$ 21.3K \\
TD3 & None & -694.8K $\pm$ 31.3K & -690.7K $\pm$ 27.6K & -694.2K $\pm$ 26.7K \\
\midrule
\multicolumn{5}{c}{\textbf{\(\sigma_{\text{test}}=1.5\)}} \\
\midrule
Algorithm & Embedding & \(\sigma_{\text{train}}=1.0\) & \(\sigma_{\text{train}}=1.5\) & \(\sigma_{\text{train}}=2.0\) \\
\midrule
SAC & Full & -496.3K $\pm$ 13.7K & -479.4K $\pm$ 16.5K & \textbf{-436.7K $\pm$ 8.8K} \\
SAC & One-hot & \textbf{-474.2K $\pm$ 10.9K} & \textbf{-475.8K $\pm$ 14.4K} & -486.0K $\pm$ 13.6K \\
SAC & None & -478.0K $\pm$ 11.7K & -483.5K $\pm$ 17.3K & -517.2K $\pm$ 11.5K \\
DDPG & Full & -992.2K $\pm$ 41.5K & -981.5K $\pm$ 42.7K & -989.9K $\pm$ 46.1K \\
DDPG & One-hot & -995.2K $\pm$ 45.2K & -952.5K $\pm$ 38.3K & -992.6K $\pm$ 27.3K \\
DDPG & None & -996.3K $\pm$ 31.6K & -1003.2K $\pm$ 42.7K & -990.6K $\pm$ 34.4K \\
TD3 & Full & -994.4K $\pm$ 43.1K & -634.7K $\pm$ 15.4K & -983.0K $\pm$ 44.1K \\
TD3 & One-hot & -959.6K $\pm$ 49.7K & -979.9K $\pm$ 32.3K & -979.8K $\pm$ 38.5K \\
TD3 & None & -989.5K $\pm$ 49.9K & -991.9K $\pm$ 64.9K & -982.8K $\pm$ 30.4K \\
\midrule
\multicolumn{5}{c}{\textbf{\(\sigma_{\text{test}}=2.0\)}} \\
\midrule
Algorithm & Embedding & \(\sigma_{\text{train}}=1.0\) & \(\sigma_{\text{train}}=1.5\) & \(\sigma_{\text{train}}=2.0\) \\
\midrule
SAC & Full & -622.7K $\pm$ 12.5K & -608.4K $\pm$ 21.1K & \textbf{-554.2K $\pm$ 12.5K} \\
SAC & One-hot & \textbf{-585.7K $\pm$ 19.2K} & \textbf{-594.0K $\pm$ 17.1K} & -617.2K $\pm$ 19.6K \\
SAC & None & -602.5K $\pm$ 15.7K & -611.3K $\pm$ 15.4K & -651.6K $\pm$ 17.9K \\
DDPG & Full & -1251.4K $\pm$ 57.0K & -1255.4K $\pm$ 53.1K & -1257.9K $\pm$ 47.5K \\
DDPG & One-hot & -1259.6K $\pm$ 55.7K & -1239.6K $\pm$ 54.7K & -1227.3K $\pm$ 54.0K \\
DDPG & None & -1253.1K $\pm$ 57.2K & -1260.0K $\pm$ 62.1K & -1243.9K $\pm$ 49.7K \\
TD3 & Full & -1245.4K $\pm$ 56.2K & -780.2K $\pm$ 16.8K & -1268.2K $\pm$ 45.9K \\
TD3 & One-hot & -1243.1K $\pm$ 49.0K & -1247.6K $\pm$ 51.4K & -1253.5K $\pm$ 29.4K \\
TD3 & None & -1227.5K $\pm$ 45.7K & -1226.1K $\pm$ 46.8K & -1219.3K $\pm$ 31.5K \\
\bottomrule
\end{tabular}}%
}
\end{table*}

\revise{Combining the training dynamics from Figure~\ref{fig:performance_comparison} with the cross-variance robustness results in Table~\ref{tab:crosssigma}, several critical observations emerge regarding algorithm performance under stochastic transit conditions:}

\paragraph{\del{(1) Convergence Speed}}
{\del{Embedding-sac rapidly achieves stable high reward within approximately 30 episodes. In contrast, maddpg-PS and MADDPG-NPS require over 150 and 180 episodes, respectively, to reach convergence. This faster ascent is accompanied by significantly reduced variance, suggesting stable and sample-efficient learning.}}
\revise{\textbf{(1) SAC's consistent superiority across all noise levels.}
SAC outperforms both TD3 and DDPG by a substantial margin under all testing conditions. At $\sigma_{\text{test}}=1.0$, SAC achieves rewards in the range of $-322$K to $-366$K, while TD3 and DDPG stagnate at approximately $-690$K to $-721$K performance that is effectively halved. This performance gap widens dramatically as environmental stochasticity increases: at $\sigma_{\text{test}}=2.0$, SAC maintains rewards around $-554$K to $-651$K, whereas TD3 and DDPG completely fail to converge, yielding catastrophic rewards near $-1,220$K to $-1,270$K. This demonstrates SAC's fundamental advantage in handling the high-dimensional, variance-sensitive state space inherent to bus fleet control.}

\paragraph{\del{(2) Asymptotic Performance}}
{\del{Embedding-SAC consistently maintains the highest mean reward throughout training, plateauing near $\sim\!-430$k. MADDPG-PS stabilizes around $\sim\!-530$k and MADDPG-NPS around $\sim\!-580$k. The uncontrolled baseline (gray dashed line) sits near $\sim\!-980$k, indicating that all methods improve upon the no-control scenario, but SAC achieves a substantially larger reward margin.}}
\revise{\textbf{(2) Catastrophic failure of TD3 and DDPG under high variance.}
Table~\ref{tab:crosssigma} reveals a striking pattern: as $\sigma_{\text{test}}$ increases from 1.0 to 2.0, TD3 and DDPG exhibit near-total collapse. Their mean rewards degrade by 3--4 times compared to low-noise conditions, with DDPG performance dropping from approximately $-700$K at $\sigma_{\text{test}}=1.0$ to below $-1,250$K at $\sigma_{\text{test}}=2.0$. TD3 displays particularly erratic behavior, with anomalous spikes (e.g., $-492$K under certain $\sigma_{\text{train}}=1.5$ configurations for $\sigma_{\text{test}}=1.0$) suggesting training instability and poor generalization. In contrast, SAC's performance degrades gracefully from approximately $-340$K to $-600$K across the same variance range indicating robust policy learning that is not overfitted to specific noise realizations.}

\paragraph{\del{(3) Training Stability}}
{\del{Shaded regions around each curve reflect the standard deviation of cumulative rewards from 15 independent evaluation episodes. Embedding-SAC exhibits the narrowest bands, especially in later stages, confirming strong policy robustness against environmental randomness. MADDPG variants suffer from wider fluctuations, with MADDPG-NPS showing instability and degradation after mid-training.}}
\revise{\textbf{(3) Full embedding architecture achieves best results under high stochasticity.}
Examining SAC variants across embedding configurations, the \emph{Full} embedding (incorporating vehicle ID, station ID, time period, and direction) consistently delivers the best performance at higher noise levels: $-322.8$K at $\sigma_{\text{test}}=1.0$, $-436.7$K at $\sigma_{\text{test}}=1.5$, and $-554.2$K at $\sigma_{\text{test}}=2.0$. This pattern confirms that rich categorical representations are essential for capturing the heterogeneity and temporal dynamics of bus operations. Simpler representations (\emph{One-hot} and \emph{None}) perform adequately under low noise but lack the representational capacity to maintain robustness as variance increases, highlighting the importance of learned embeddings in stochastic environments.}

\paragraph{\del{(4) parameter sharing effects}}
{\del{MADDPG-PS benefits from shared parameters, leading to faster early-stage improvement compared to MADDPG-NPS. However, its inability to capture agent heterogeneity under bidirectional and timetable-driven deployment prevents it from achieving SAC-level rewards.}}
\revise{\textbf{(4) Rapid convergence and superior sample efficiency.}
Figure~\ref{fig:performance_comparison} illustrates that SAC converges to stable high rewards within approximately 30 episodes, while MADDPG-PS and MADDPG-NPS require over 150 and 180 episodes, respectively. SAC's plateau near $\sim\!-430$K contrasts sharply with MADDPG-PS at $\sim\!-530$K and MADDPG-NPS at $\sim\!-580$K, all far superior to the uncontrolled baseline near $\sim\!-980$K. Moreover, the shaded variance bands around SAC remain narrow throughout training, confirming both fast convergence and low variability critical for real-world deployment where sample efficiency and reliability are paramount.}

\paragraph{\del{(5) Variance-Sensitive Architecture}}
{\del{Embedding-SAC's success lies in its single-agent architecture enhanced with categorical embeddings for vehicle ID, station ID, time period, and direction. This encoding strategy compensates for uneven exposure to training samples, particularly during peak-hour imbalances, and facilitates generalization across dynamic spatiotemporal contexts.}}

\paragraph{\del{(6) Rationale for Not Using Waiting Time}}
{\del{Passenger waiting time, though common in public transit metrics, is excluded here. As emphasized by Ceder~\cite{ceder2016public}, waiting time is more influenced by the timetable-demand alignment than by holding interventions. Since our method focuses on intra-trip headway stabilization, cumulative reward and headway-based variance are more indicative of control quality.}}

\paragraph{\del{(7) Realistic Data Generation}}
{\del{Simulated demand and road conditions are derived from real-world patterns. OD-based passenger arrivals follow Poisson processes updated hourly, while segment-wise travel speeds fluctuate according to time-varying Gaussian distributions, consistent with~\cite{zheng2023robust}.}}

\vspace{0.5em}
\noindent
{\del{\textbf{Conclusion:} Embedding-SAC outperforms all MADDPG baselines and the no-control benchmark by a large margin. Its reward gain, stability, and convergence efficiency highlight the strength of combining soft actor-critic methods with categorical feature embedding in realistic, stochastic transit environments.}}

\revise{\textbf{Summary:} The combined evidence from training curves and cross-variance evaluation demonstrates that SAC with full categorical embeddings is uniquely suited for robust bus fleet control. While TD3 and DDPG completely fail under realistic noise levels ($\sigma \geq 1.5$), SAC maintains both high performance and stability, converging 5--6 times faster than multi-agent alternatives. This superiority stems from SAC's entropy-regularized objective and its ability to leverage rich learned representations of vehicle identity, spatiotemporal context, and directional heterogeneity capabilities essential for real-world transit systems operating under unpredictable demand and traffic conditions.}

\revise{
To further assess robustness and the contribution of categorical embeddings, we conduct a full ablation study across noise levels $\sigma\!\in\!\{1.0,1.5,2.0\}$ and three input representations: 
(i)~\emph{No categorical features} (IDs removed), 
(ii)~\emph{One-hot encoding}, and 
(iii)~\emph{Learnable embeddings}. 
Each model (SAC, TD3, DDPG) is trained under every $\sigma_{\mathrm{train}}$ and evaluated under each $\sigma_{\mathrm{test}}$, producing a total of $3{\times}3{\times}3{=}27$ combinations.
Table~\ref{tab:crosssigma} summarizes the best results for each testing variance (complete results are provided in the supplementary CSV \texttt{cross\_sigma\_all\_results.csv}; the subset of best scores is stored in \texttt{cross\_sigma\_best\_results.csv}).}

\revise{
Several critical insights emerge from Table~\ref{tab:crosssigma}:
(1)~\textbf{SAC dominates across all noise levels.} Under every $\sigma_{\text{test}}$ condition, SAC variants consistently outperform TD3 and DDPG by factors of 2--4$\times$. Even SAC's worst configuration (None embedding at $\sigma_{\text{test}}=2.0$: $-651.6$K) substantially exceeds the best TD3 or DDPG results at the same noise level (approximately $-1,220$K to $-1,270$K).
(2)~\textbf{TD3 and DDPG fail catastrophically under high variance.} At $\sigma_{\text{test}}=1.5$ and $2.0$, both algorithms exhibit near-complete collapse, with rewards degrading to $-950$K to $-1,270$K worse than 3$\times$ their low-noise performance. This suggests fundamental brittleness in deterministic policy gradient methods when applied to high-dimensional, stochastic transit control.
(3)~\textbf{Full embedding excels under stochasticity.} SAC with Full embedding achieves the best reward in every $\sigma_{\text{test}}$ setting (bolded in table): $-322.8$K, $-436.7$K, and $-554.2$K for $\sigma_{\text{test}}=1.0, 1.5, 2.0$ respectively. In contrast, simpler representations (One-hot, None) show diminished robustness as noise increases, confirming that learned categorical embeddings are essential for generalizing across diverse operational conditions.
(4)~Cross-variance generalization is algorithm-dependent. SAC's performance degrades smoothly from $-322$K to $-554$K as $\sigma_{\text{test}}$ doubles, while TD3 and DDPG exhibit erratic, non-monotonic behavior (e.g., TD3 Full's anomalous $-492$K and $-634$K spikes), indicating poor policy stability and overfitting to specific noise realizations.
Overall, these results demonstrate that SAC with rich categorical embeddings uniquely combines high performance, robustness to variance shifts, and stable cross-domain generalization properties critical for real-world transit systems where demand and travel times vary unpredictably.}

\subsection{Empirical evidence of bus bunching and control effectiveness}

To evaluate the effectiveness of our proposed SAC-based holding strategy and to visually demonstrate how bus bunching occurs and propagates, we present a series of trajectory-based and statistical analyses.

\begin{figure}[htbp]
\centering
\includegraphics[width=\textwidth]{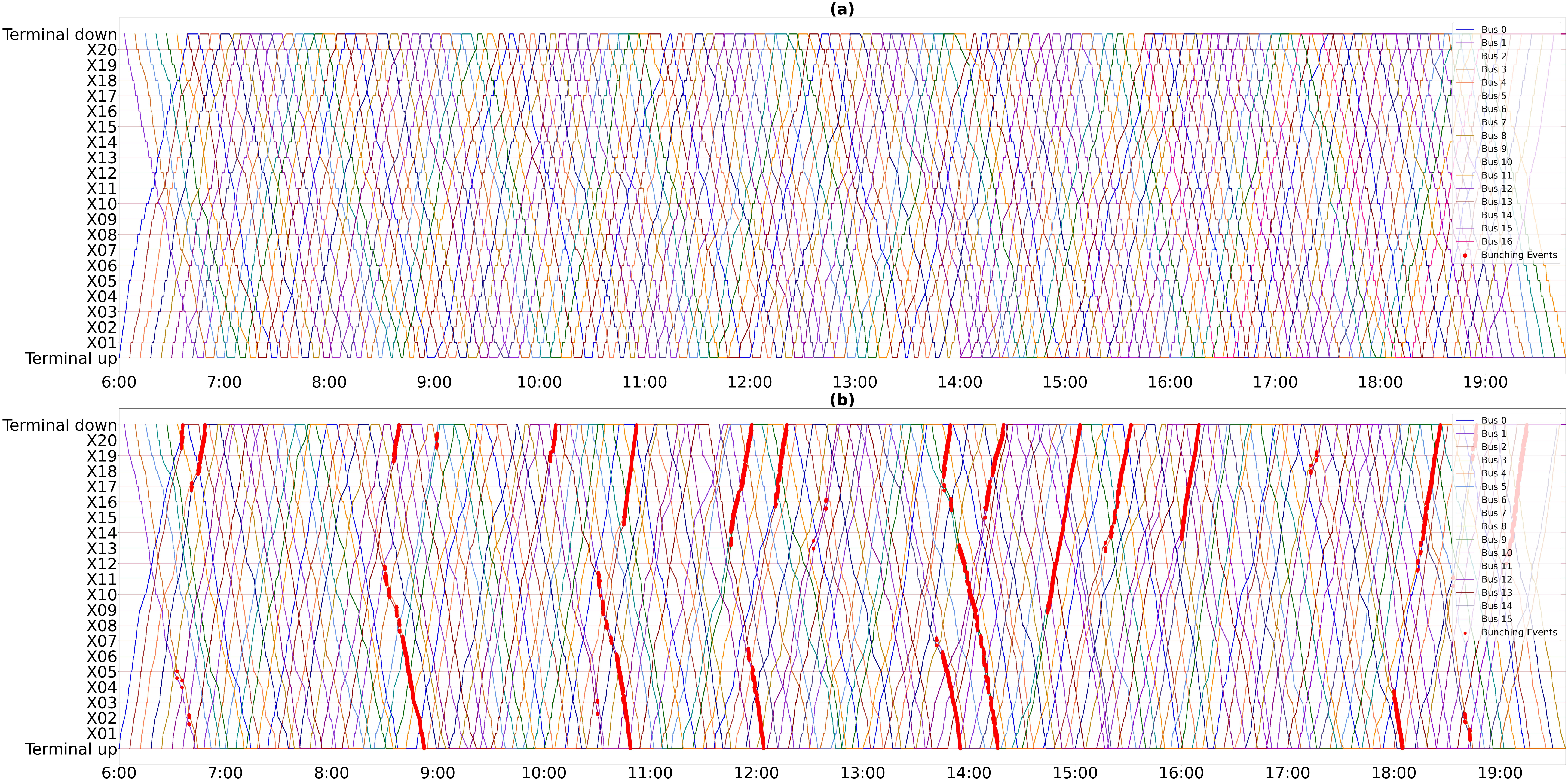}
\caption{Bus trajectory visualization with and without SAC-based control. Top: (a) Bus trajectories with SAC control. Bottom: (b) Bus trajectories without control. Red markers indicate detected bunching events.}
\label{fig:bunching_control_trajectory}
\end{figure}

Figure~\ref{fig:bunching_control_trajectory} visualizes the complete operational period (06:00--19:00) for both the controlled and uncontrolled scenarios. Each colored line depicts the trajectory of an individual bus as it travels between terminals, traversing all 22 stops. In the uncontrolled scenario (b, bottom panel), frequent bunching events are observed highlighted by \textit{\textcolor{red}{red}} segments indicating intervals where two or more buses operate in close proximity. These bunching occurrences are most prevalent during the early morning (06:30--08:30) and late afternoon (16:00--18:00) periods, immediately preceding the peak demand hours.

In contrast, the controlled scenario (a, top panel), where SAC regulates holding decisions at stops, demonstrates a complete absence of bunching events throughout the episode. Bus trajectories remain uniformly spaced, confirming the robustness and effectiveness of the proposed control strategy under stochastic demand and travel speed conditions.

\begin{figure}[htbp]
\centering
\includegraphics[trim = 20 20 20 10, width=0.80\textwidth]{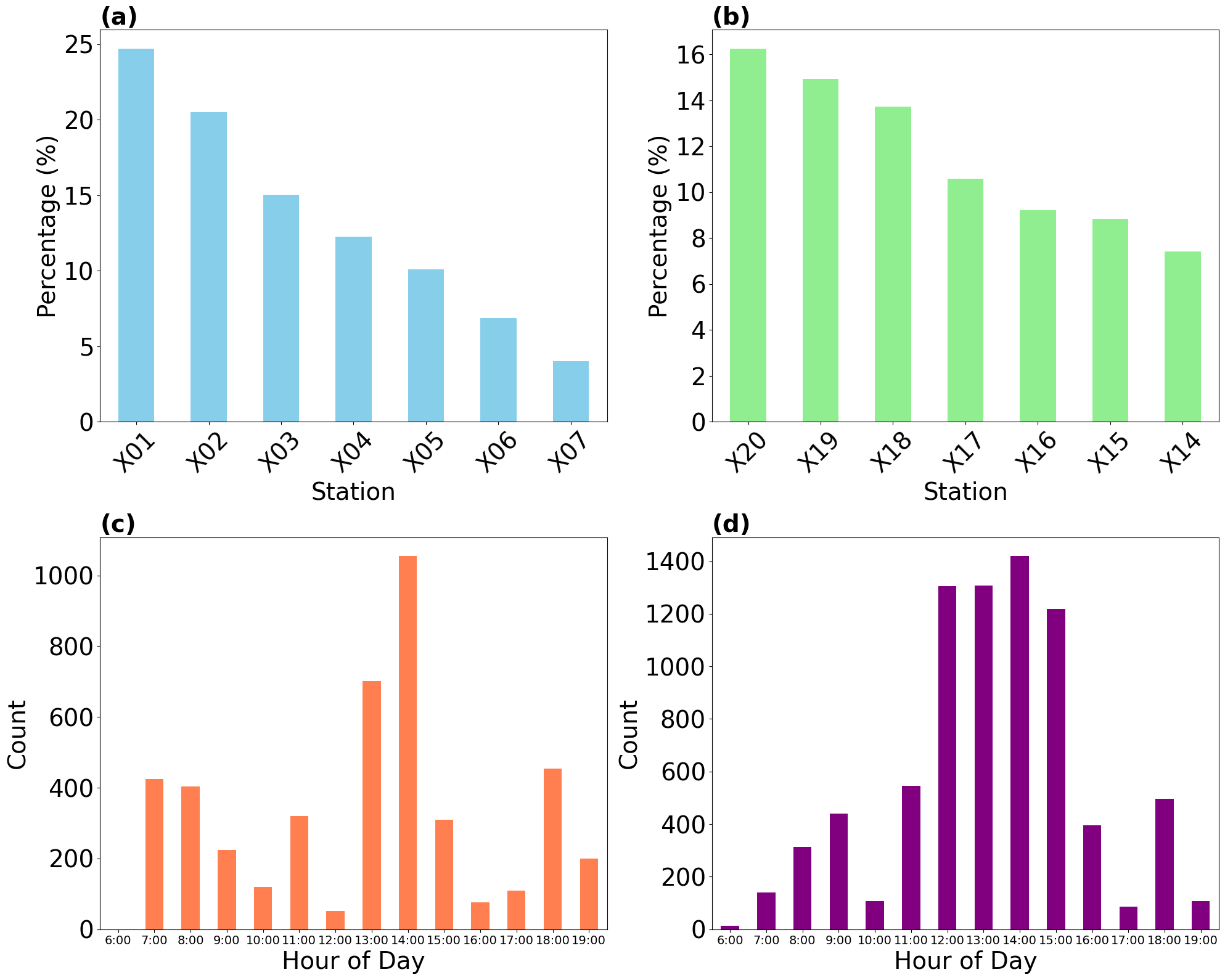}
\caption{Spatio-temporal analysis of bunching events. (a) Top 7 bunching-prone stations (down direction). (b) Top 7 bunching-prone stations (up direction). (c) Hourly bunching frequency (down direction). (d) Hourly bunching frequency (up direction).}
\label{fig:bunching_stats}
\end{figure}

Figure~\ref{fig:bunching_stats} provides a fine-grained analysis of the spatio-temporal distribution of bunching events observed in the uncontrolled case. Subplots (a) and (b) identify the top 7 stops most frequently involved in bunching, stratified by travel direction. These are predominantly downstream stops near the end of each route, where accumulated delays due to traffic precipitate bus convergence.
Subplots (c) and (d) depict hourly bunching frequencies in both directions. The risk of bunching is most pronounced during transitional hours especially 07:00--09:00 and 16:00--18:00 which precede the peak intensity of rush hour demand. This corroborates our earlier hypothesis that bunching is not exclusively a peak-hour phenomenon, but rather the result of demand-speed imbalances accumulating in the periods preceding peak load.

Together, these findings reinforce the key design motivations of our work:

\begin{itemize}
    \item Bunching tends to emerge in pre-peak and transitional periods due to localized demand or speed perturbations;
    \item Spatially, bunching intensifies near the end of each trip when the bus has been exposed to long sequences of stochastic disturbances;
    \item A control strategy that integrates spatio-temporal context facilitated by SAC with categorical embedding can effectively eliminate bunching under significant stochasticity.
\end{itemize}

These results validate the proposed modeling and control approach, and demonstrating that a robustly trained single-agent RL policy can generalize across dynamic fleet states and operational conditions.

\section{Conclusion}

This paper presents a novel single-agent reinforcement learning framework for mitigating bus bunching in realistic, bidirectional, timetabled transit systems. Departing from conventional multi-agent approaches designed for idealized loop-line environments, our method explicitly incorporates categorical embeddings—such as vehicle ID, station ID, and trip identifiers into the SAC policy. This enables a single-agent to generalize across heterogeneous agents and spatio-temporal contexts, overcoming the limitations of data imbalance and agent-specific training instability inherent in traditional MARL settings.

To emulate operational realism, we construct a bidirectional bus simulation environment based on empirically derived passenger demand and route speed profiles. The environment features a timetable-driven dispatch mechanism and stochastic variability in both road conditions and boarding/alighting dynamics. This setup, extending and refining the methodology proposed by Zheng et al.~\cite{Zheng2024RobustNonlinearBusSpeedControl}---supports more faithful modeling of real-world operational challenges.

Through extensive experiments, we demonstrate that Embedding-SAC achieves better performance in terms of cumulative mean reward value, asymptotic stability, and variance of cumulative reward value compared to both parameter-sharing and non-sharing variants of MADDPG. Notably, our control policy eliminates all bunching events across the entire simulation period, as confirmed by trajectory-level visualizations and spatio-temporal heatmaps. These results validate that a well-structured single-agent policy, when leveraging appropriate embedded features, can effectively optimize holding decisions under uncertainty.

In future work, we plan to extend this framework to incorporate non-stationary demand distributions, transfer learning across corridors, and integration with high-level scheduling modules. We also intend to explore hierarchical and latent-graph models to further disentangle causal mechanisms in bus dynamics.

\section*{Acknowledgments}
This work was supported by the National Natural Science Foundation of China (Grant No. 72371251), the Natural Science Foundation for Distinguished Young Scholars of Hunan Province (Grant No. 2024JJ2080), and the Key Research and Development Program of Hunan Province of China (Grant No. 2024JK2007).

\bibliography{bus_control_refs}

@article{Nie2024MultiStrategy,
  author  = {{Nie Qinghui} and {Ou Jishun} and {Zhang Haiyang} and {Lu Jiawei} and {Li Shen} and {Shi Haotian}},
  title   = {A robust integrated multi-strategy bus control system via deep reinforcement learning},
  journal = {Engineering Applications of Artificial Intelligence},
  volume  = {133},
  pages   = {107986},
  year    = {2024}
}

@article{Tang2024SmartCities,
  author  = {{Tang Yuhan} and {Qu Ao} and {Jiang Xuan} and {Mo Baichuan} and {Cao Shangqing} and {Rodriguez Joseph} and {Koutsopoulos Haris N.} and {Wu Cathy} and {Zhao Jinhua}},
  title   = {Robust Reinforcement Learning Strategies with Evolving Curriculum for Efficient Bus Operations in Smart Cities},
  journal = {Smart Cities},
  volume  = {7},
  number  = {6},
  pages   = {3658--3677},
  year    = {2024}
}

@article{Yu2024HMARL,
  author  = {{Yu Mengdi} and {Yang Tao} and {Li Chunxiao} and {Jin Yaohui} and {Xu Yanyan}},
  title   = {Mitigating Bus Bunching via Hierarchical Multi-Agent Reinforcement Learning},
  journal = {IEEE Transactions on Intelligent Transportation Systems},
  volume  = {25},
  number  = {8},
  pages   = {9675--9680},
  year    = {2024}
}

@inproceedings{Eysenbach2022MaxEnt,
  author    = {{Eysenbach Benjamin} and {Levine Sergey}},
  title     = {Maximum Entropy RL (Provably) Solves Some Robust RL Problems},
  booktitle = {International Conference on Learning Representations (ICLR)},
  year      = {2022}
}

@inproceedings{Menda2019EventDriven,
  author    = {{Menda Kunal} and {Chen Yi-Chun} and {Grana Justin} and {Bono James W.} and {Tracey Brendan D.} and {Kochenderfer Mykel J.} and {Wolpert David}},
  title     = {Deep Reinforcement Learning for Event-Driven Multi-Agent Decision Processes},
  journal   = {IEEE Transactions on Intelligent Transportation Systems},
  year      = {2019},
  volume={20},
  number={4},
  pages={1259-1268},
}

@book{ceder2016public,
  author    = {{Ceder Avishai}},
  title     = {Public Transit Planning and Operation: Modeling, Practice and Behavior},
  publisher = {CRC Press},
  year      = {2016},
  edition   = {Second}
}

@article{cats2012bus,
  author  = {{Cats Oded} and {Nabavi Larijani Anahid} and {Olafsdottir Asdis} and {Burghout Wilco} and {Andreasson Ingmar} and {Koutsopoulos Haris N.}},
  title   = {Bus-holding control strategies: Simulation-based evaluation and guidelines for implementation},
  journal = {Transportation Research Record},
  volume  = {2274},
  number  = {1},
  pages   = {100--108},
  year    = {2012}
}

@article{bie2020dynamic,
  author  = {{Bie Yiming} and {Xiong Xinyu} and {Yan Yadan} and {Qu Xiaobo}},
  title   = {Dynamic headway control for high-frequency bus lines based on speed guidance and intersection signal adjustment},
  journal = {Computer-Aided Civil and Infrastructure Engineering},
  volume  = {35},
  number  = {1},
  pages   = {4--25},
  year    = {2020}
}

@article{cats2019frequency,
  author  = {{Cats Oded} and {Glück Stefan}},
  title   = {Frequency and vehicle capacity determination using a dynamic transit assignment model},
  journal = {Transportation Research Record: Journal of the Transportation Research Board},
  volume  = {2673(3)},
  pages   = {574--585},
  year    = {2019}
}

@article{Zheng2024RobustNonlinearBusSpeedControl,
  title     = {Robust nonlinear decision mapping approach for online bus speed control under uncertainty},
  author    = {{Zheng Liang} and {Liu Pengjie}},
  journal   = {Computer-Aided Civil and Infrastructure Engineering},
  volume    = {39},
  number    = {2},
  pages     = {203--221},
  year      = {2024},
  publisher = {Wiley},
  doi       = {10.1111/mice.13064}
}

@article{cortes2010hybrid,
  author  = {{Cortés Cristián E.} and {Sáez Doris} and {Milla Freddy} and {Núñez Alfredo} and {Riquelme Marcela}},
  title   = {Hybrid predictive control for real-time optimization of public transport systems' operations based on evolutionary multi-objective optimization},
  journal = {Transportation Research Part C: Emerging Technologies},
  volume  = {18},
  number  = {5},
  pages   = {757--769},
  year    = {2010}
}

@article{haarnoja2018soft,
  author  = {{Haarnoja Tuomas} and {Zhou Aurick} and {Abbeel Pieter} and {Levine Sergey}},
  title   = {Soft Actor-Critic: Off-policy maximum entropy deep reinforcement learning with a stochastic actor},
  journal = {arXiv preprint arXiv:1801.01290},
  year    = {2018}
}

@inproceedings{wang2021reducing,
  author    = {{Wang Jiawei} and {Sun Lijun}},
  title     = {Reducing bus bunching with asynchronous multi-agent reinforcement learning},
  booktitle = {Proceedings of the Thirtieth International Joint Conference on Artificial Intelligence (IJCAI-21)},
  year      = {2021}
}

@inproceedings{haarnoja2018applications,
  title     = {Soft Actor-Critic Algorithms and Applications},
  author    = {{Haarnoja Tuomas} and {Zhou Aurick} and {Hartikainen Kristian} and {Tucker George} and {Ha Sehoon} and {Tan Jie} and {Kumar Vikash} and {Zhu Henry} and {Gupta Abhishek} and {Abbeel Pieter} and {Levine Sergey}},
  booktitle = {Proceedings of the 37th International Conference on Machine Learning (ICML)},
  year      = {2020},
  series    = {PMLR},
  volume    = {119},
  pages     = {4912--4922}
}

@article{Zheng2022EJOR,
  author  = {{Zheng Liang} and {Bao Ji} and {Xu Chengcheng.} and {Tan Zhen}},
  title   = {Biobjective robust simulation-based optimization for unconstrained problems},
  journal = {European Journal of Operational Research},
  journal = {European Journal of Operational Research},
  volume = {299},
  number = {1},
  pages = {249-262},
  year = {2022},
  issn = {0377-2217}
}

@article{wang2020dynamic,
  title     = {Dynamic holding control to avoid bus bunching: A multi-agent deep reinforcement learning framework},
  author    = {{Wang Jiawei} and {Sun Lijun}},
  journal   = {Transportation Research Part C: Emerging Technologies},
  volume    = {116},
  pages     = {102661},
  year      = {2020},
  publisher = {Elsevier}
}

@article{Daganzo2011cooperation,
  title     = {Reducing bunching with bus-to-bus cooperation},
  author    = {{Daganzo Carlos F.} and {Pilachowski Josh}},
  journal   = {Transportation Research Part B: Methodological},
  volume    = {45},
  number    = {2},
  pages     = {267--277},
  year      = {2011},
  publisher = {Elsevier}
}

@article{REZAZADA2024766,
  title    = {Bus bunching: a comprehensive review from demand, supply, and decision-making perspectives},
  journal  = {Transport Reviews},
  volume   = {44},
  number   = {4},
  pages    = {766-790},
  year     = {2024},
  issn     = {0144-1647},
  author   = {{Rezazada Mustafa} and {Nassir Neema} and {Tanin Egemen} and {Ceder Avishai}}
}

@article{wu2017modelling,
  title     = {Modelling bus bunching and holding control with vehicle overtaking and distributed passenger boarding behaviour},
  author    = {{Wu Weitiao} and {Liu Ronghui} and {Jin Wenzhou}},
  journal   = {Transportation Research Part B: Methodological},
  volume    = {104},
  pages     = {175--197},
  year      = {2017},
  publisher = {Elsevier}
}

@misc{bunching_wiki,
  title        = {Bus Bunching - Wikipedia},
  howpublished = {\url{https://en.wikipedia.org/wiki/Bus_bunching}},
  note         = {Accessed: 2025-07-29}
}

@article{Tang2023TRD,
  title     = {Optimization of single-line electric bus scheduling with skip-stop operation},
  author    = {{Tang Chunyan} and {Shi Hudi} and {Liu Tao}},
  journal   = {Transportation Research Part D: Transport and Environment},
  volume    = {117},
  pages     = {103652},
  year      = {2023},
  publisher = {Elsevier}
}

@article{Tang2023Transportmetrica,
  title     = {Optimisation of a new hybrid transit service with modular autonomous vehicles},
  author    = {{Tang Chunyan} and {Liu Jinqiang} and {Ceder Avishai} and {Jiang Yu}},
  journal   = {Transportmetrica A: Transport Science},
  year      = {2024},
  volume = {20},
  number = {2},
  publisher = {Taylor \& Francis}
}

@article{Tang2023IJST,
  title     = {Optimal selection of vehicle types for an electric bus route with shifting departure times},
  author    = {{Tang Chunyan} and {Ge Ying-En} and {Xue He} and {Ceder Avishai} and {Wang Xiaokun}},
  journal   = {International Journal of Sustainable Transportation},
  volume = {17},
  number = {11},
  pages = {1217--1235},
  year = {2023},
  publisher = {Taylor \& Francis},
}

@article{Tang2021TRR,
  title     = {Public Transport Fleet Replacement Optimization Using Multi-Type Battery-Powered Electric Buses},
  author    = {{Tang Chunyan} and {Li Xiaoyu} and {Ceder Avishai} and {Wang Xiaokun}},
  journal   = {Transportation Research Record},
  volume    = {2675},
  number    = {12},
  pages     = {1422--1431},
  year      = {2021},
  publisher = {SAGE}
}

@article{Tang2020IJST,
  title     = {Optimal operational strategies for single bus lines using network-based method},
  author    = {{Tang Chunyan} and {Ceder Avishai} and {Ge Ying-En} and {Liu Tao}},
  journal   = {International Journal of Sustainable Transportation},
  volume = {15},
  number = {5},
  pages = {325--337},
  year = {2021},
  publisher = {Taylor \& Francis}
}

@article{Zheng2024AOR,
  title     = {Robust Simulation-based Optimization for Multiobjective Problems with Constraints},
  author    = {{Zheng Liang} and {Bao Ji} and {Tan Zhen}},
  journal   = {Annals of Operations Research},
  volume    = {346},
  number    = {2},
  pages     = {1897--1927},
  year      = {2024},
  publisher = {Springer}
}

@article{Liu2024RobustBus,
  author  = {{Liu Pengjie} and {Zheng Liang} and {Zheng Nan}},
  title   = {Bi-objective robust nonlinear decision approach for en-route bus speed control considering implementation errors and traffic uncertainties},
  journal = {Transportation Research Part C: Emerging Technologies},
  volume  = {169},
  pages   = {104870},
  year    = {2024}
}
\bibliographystyle{IEEEtran}
\end{document}